\documentclass[conference]{IEEEtran}
\usepackage{amsmath,amssymb}
\usepackage{algorithm}
\usepackage{algpseudocode}
\usepackage{graphicx}
\usepackage[caption=false]{subfig}
\usepackage{algorithm}
\usepackage{caption}

\PassOptionsToPackage{noend}{algpseudocode}
\usepackage{algpseudocode}

\errorcontextlines\maxdimen

\makeatletter
\newcommand*{\algrule}[1][\algorithmicindent]{\makebox[#1][l]{\hspace*{.5em}\vrule height .75\baselineskip depth .25\baselineskip}}%

\newcount\ALG@printindent@tempcnta
\def\ALG@printindent{%
    \ifnum \theALG@nested>0
        \ifx\ALG@text\ALG@x@notext
            \addvspace{-3pt}
        \else
            \unskip
            \ALG@printindent@tempcnta=1
            \loop
                \algrule[\csname ALG@ind@\the\ALG@printindent@tempcnta\endcsname]%
                \advance \ALG@printindent@tempcnta 1
            \ifnum \ALG@printindent@tempcnta<\numexpr\theALG@nested+1\relax
            \repeat
        \fi
    \fi
    }%
\usepackage{etoolbox}
\patchcmd{\ALG@doentity}{\noindent\hskip\ALG@tlm}{\ALG@printindent}{}{\errmessage{failed to patch}}
\makeatother
\ifCLASSINFOpdf
\else
\fi
\begin{document}
%
\title{Intelligent Device Discovery in the Internet of Things -- Enabling the Robot Society}


\author{\IEEEauthorblockN{James Sunthonlap, Phuoc Nguyen, Zilong Ye}
\IEEEauthorblockA{California State University Los Angeles \\ 5151 State University Drive, Los Angeles, CA 90032, USA\\
Email: zye5@calstatela.edu}
}



%


\maketitle

\begin{abstract}
The Internet of Things (IoT) is continuously growing to connect billions of smart devices anywhere and anytime in an Internet-like structure, which enables a variety of applications, services and interactions between human and objects. 
In the future, the smart devices are supposed to be able to autonomously discover a target device with desired features and thus yield the computing service, network service and data fusion that leads to the generation of a set of entirely new services and applications that are not supervised or even imagined by human beings.
The pervasiveness of smart devices, as well as the heterogeneity of their design and functionalities, raise a major concern: How can a smart device efficiently discover a desired target device? 
In this paper, we propose a \emph{Social-Aware and Distributed} (SAND) scheme that achieves a fast, scalable and efficient device discovery in the IoT. 
The proposed SAND scheme adopts a novel device ranking criteria that measures the device's degree, social relationship diversity, clustering coefficient and betweenness.
Based on the device ranking criteria, the discovery request can be guided to travel through critical devices that stand at the major intersections of the network, and thus quickly reach the desired target device by contacting only a limited number of intermediate devices.
We conduct comprehensive simulations on both random networks and scale-free networks to evaluate the performance of SAND in terms of the discovery success rate, the number of devices contacted and the number of communication hops. 
The simulation results demonstrate the effectiveness of SAND.
With the help of such an intelligent device discovery as SAND, the IoT devices, as well as other computing facilities, software and data on the Internet, can autonomously establish new social connections with each other as human being do. 
They can formulate self-organized computing groups to perform required computing tasks, facilitate a fusion of a variety of computing service, network service and data to generate novel applications and services, evolve from the individual aritificial intelligence to the collaborative intelligence, and eventually enable the birth of a robot society.
\end{abstract}


%
\IEEEpeerreviewmaketitle

\section{Introduction}
The Internet of Things (IoT) is continuously growing to connect billions of smart devices anywhere and anytime through an Internet-like structure. A recent forecast by International Data Corporation (IDC) shows that the IoT will involve more than 212 billion of objects by 2020, and the IoT and the associated ecosystem are predicted to have a \$1.7 trillion market \cite{forecast}. The IoT devices are capable of sensing, analyzing and evaluating the surrounding objects and people. They can collaborate and work together to provide a set of new applications and services, such as smart home, e-healthcare and intelligent transportation system. The IoT is envisioned to dramatically change and enhance the interactions between human and objects, bringing transformative benefits to the lives of human beings.

As the IoT evolves, it is promising that the future smart devices may have the capability to discover a target device with desired features, and autonomously collaborate with each other to accomplish certain missions or tasks.
An intelligent device discovery strategy may allow the IoT devices to efficiently establish new connections with other devices based on their need. The connections can be set up to gather a certain number of computing powers, network functions, program source codes, raw datasets, and etc to enable new services and applications. A variety of computing service, network service and data can be fused \cite{sun1}-\cite{chang1} by following the guidance of an intelligent device discovery strategy. For example, a glucose level monitor can send a request to find a glucose analyzer, and collaborate on evaluating a given patient's glucose level. The glucose analyzer may also need to search for a national wide database to compare the given patient's results with that of other patients, and then given the advisement or alert to the given patient. Here, an intelligent device discovery strategy could guide the request to the appropriate destinations for obtaining the required computing facilities, functions and data to provide the evaluation service,  natually leading to a computing service, network service and data fusion.
In addition, the IoT devices can also form social connections as human beings do, and they can collaborate to generate entirely new kinds of applications and services with the help of collaborative intelligence of smart devices and data, rather than the supervision of human beings.
In the above future scenarios, one of the key challenges is how to achieve a fast, scalable and efficient device discovery when there are millions or billions of devices in the IoT. 
An intelligent device discovery should have a forwarding strategy that guide the discovery request to quickly arrive at the desired target device with minimum detours. In addition, since most of the IoT devices are power constrained, the device discovery should avoid involving too many intermediate devices in the process of data exchange. 

There are a few existing works on the topic of IoT device discovery. The work in \cite{barnaghi} identified the differences between user search and machine-to-machine search, and presented the challenges and requirements for IoT device search. The authors in \cite{ccori} studied device discovery on different network topologies, including a star-like centralized network, a regular mesh-like decentralized network and a hierarchical network. They focused on analyzing the effect of the network topology on the discovery success rate, rather than designing new device discovery strategies. The study in \cite{context} proposed a centralized IoT search engine that can accurately interpret the context of the discovery request and thus make a proper management of search and usage in the IoT middleware environment. However, this centralized approach may not be scalable when the network size is large, and can be vulnerable to a single point of failure. More recent work \cite{icn} investigated a distributed device discovery strategy based on Information-Centric Networking (ICN). However, this approach may not be applicable to the IoT system since it relies on the built-in features of ICN such as cache and broadcast capabilities, which could introduce an extra cost to the resource-constrained IoT devices. 

In this work, we focus on investigating autonomous and intelligent device discovery in the IoT. We do not treat smart devices as independent objects or limit our study on the artificial intelligence of a single device, but we explore human-like social behaviors and collaborative intelligence of smart devices. 
It is envisioned for the next episode of research in the IoT and robotics.
In particular, we propose a novel IoT device discovery scheme, called \emph{Social-Aware and Distributed} (SAND), which can discover a desired target device in a fast, efficient and scalable manner. In SAND, IoT devices are fair and equal as peers, and they can interact with each other in an autonomous and distributed manner as human do, as each of them only maintains the information of their local neighboring peers. 
Each IoT device is ranked by measuring the device's degree, social relationship diversity, clustering coefficient and betweenness. Based on this, SAND opts to forward the discovery request to the neighboring device with the highest rank, which is likely to stand at the major intersection of the network that can fast and efficiently lead the request to the desired target device.
We conduct comprehensive simulations to evaluate the proposed SAND scheme on both a random network and a scale-free network. The simulation results show that SAND can achieve a high discovery success rate and a short discovery path, while contacting a small number of intermediate devices.

The rest of the paper is organized as follows. We first introduce the traditional centralized and distributed device discovery schemes in Section 2. Then, we describe the design details of the proposed SAND scheme in Section 3. In Section 4, we present the performance evaluation. In Section 5, we discuss the research challenges and opportunities related to the IoT device discovery. Finally, we conclude the paper and propose our future work in Section 6.

\section{Traditional IoT Device Discovery Schemes}

There are two traditional solutions for addressing the device discovery in the IoT, which are the centralized scheme and the distributed scheme. The former leverages a centralized controller to resolve the discovery request, and the latter uses a simple broadcast mechanism to discover the required target device in a breadth-first search manner. The detailed description and the comparison of these two approaches are presented in the following subsections.

\subsection{Centralized IoT Device Discovery Scheme}
To achieve a fast and efficient IoT device discovery, a simple yet effective approach is to introduce a centralized controller which keeps track of the general information (e.g., the device's functionality) of all the devices and maintains the shortest paths between all the device pairs. If there exists a source device which requests to discover a target device that provides a desired function, the discovery request can be sent to the centralized controller to resolve. The centralized controller can find the location of the desired target device and reply to the source device with the shortest path information to the target device. Such a centralized scheme is fast and accurate due to that the centralized controller has a global view of all the devices in the network. It is also efficient since the centralized controller can provide the shortest path between the source and target devices, which involves the minimum number of intermediate devices in transmitting the discovery request.  

The centralized scheme can be designed with robustness against dynamic changes in the network. When a new device joins the network, it needs to register the general information at the controller. A device that leaves the network also needs to unsubscribe at the centralized controller. In addition, the centralized controller can periodically send out heartbeat messages to all the registered devices and maintain an up-to-date list of the alive devices in the network. If a device fails to reply to the heartbeat after three attempts, it is considered as in the mode of lost and will be removed from the list of alive at the centralized controller. 
We can see that the centralized scheme highly relies on the centralized controller which manages the whole system and the communication between the IoT devices.

\subsection{Distributed IoT Device Discovery Scheme}
The centralized scheme is fast and efficient; however, it may not be scalable when the number of devices is large. In contrast, the distributed scheme offers a scalable device discovery in the IoT. In the distributed scheme, each device maintains a local neighbor list that consists of the neighbors' general information. When a new device joins a network, it only exchanges the general information with nearby devices that are within the transmission range. When a device leaves the network, it needs to inform its neighbors for them to update their local neighbor list. Heartbeats can be exchanged between the devices to maintain the fresh live or lost status of the devices' neighbors. The distributed discovery scheme works in a breadth-first search manner. When a source device makes a request to discover a target device with desired features, the distributed scheme simply broadcasts the discovery request to all the neighboring devices. This process iterates until the desired target device is found. 

Compared to the centralized scheme, the distributed scheme is more resilient against possible failures since there is no centralized control that is at the risk of single point of failure. It is also more robust when the network is dynamically changing since each device only needs to maintain their local neighbor's information. We can also infer that the distributed scheme can find a cost-efficient communication path between the source and target devices since the breadth-first search is used. The only limitation is that a large number of intermediate devices may be involved in transmitting the discovery request when broadcast is used, which is energy consuming. This is harmful to the IoT system where most of the devices are power constrained.

\section{Social-Aware and Distributed IoT Device Discovery -- SAND}

In the famous small-world experiment \cite{travers}, Travers and Milgram found that people are tied by a short chain of connections and any two persons can be connected in six hops by simply exploring their social networks. A more recent Facebook research \cite{fb} confirmed this observation in which they concluded that, among their 1.59 billion active users, in average a person is separated by 3.57 hops away from another person. As the IoT evolves, devices could also exhibit human-like social relationships. For example, devices can be considered as family if they are from the same manufacturer, or can be considered as colleagues if they work together to provide a service. Such social aspects can therefore be used to intelligently navigate device discovery requests through the device social network and connect two devices in a few hops as the small-world exhibited by the human social network. Based on this motivation, we propose the \emph{Social-Aware and Distributed} (SAND) scheme, which can achieve a scalable, fast and efficient device discovery in the IoT. In SAND, IoT devices can autonomously establish meaningful relationships with other devices, and form an overlay device social network in addition to their communication network. It may be faster and more efficient for the devices to find their desired target device by searching among their ``friends" through the overlay device social network, as oppose to simply broadcasting through the communication network as the distributed scheme works. In the following subsections, we will present the details of the SAND architecture, the device ranking criteria and the forwarding strategy, respectively.

\subsection{The SAND Architecture}
In SAND, the IoT devices form two layers of networks, which consists of the communication network and the overlay device social network. The communication network is at the bottom layer, which abstracts the data communication links between the IoT devices. As long as two IoT devices are within their transmission range, a communication link exists between them. Actually, the distributed scheme introduced in the previous section only considers this communication network and performs simple broadcasting for device discovery. In SAND, in addition to the communication network, an overlay device social network is constructed to help achieve an effective IoT device discovery. When two IoT devices can communicate with each other, they can exchange their device general information, such as the manufacturer, the functionality, the ownership and the location information. In the overlay device social network, there exists a link between two IoT devices if they have a valid social relation. Similar as human relationships, device social relationships typically include family, friends, neighbors, colleagues, and etc. Devices from the same manufacturer can be considered as family, e.g., an iPhone and an Apple TV. Device friendship can be described as objects that interact frequently with one another and tend to share a common theme. For example, Bob's smartphone and his body sensors are considered as friends since they interact frequently and they both serve as key components in providing e-healthcare services. IoT devices that locate in the same room or floor can be considered as neighbors. IoT devices are considered as colleagues if they work together to provide a specific service. For example, the temperature sensors, the humidity sensors and the air conditioner are colleagues that work together to offer a comfortable living environment in a smart home. It is worth noting that, in our simulation (to be introduced in Section IV), we abstract the device general information into a number of device features, and there exists a social link between two IoT devices if they have a common feature.

\begin{figure}[!t]
\centering
\includegraphics[width=3.5in]{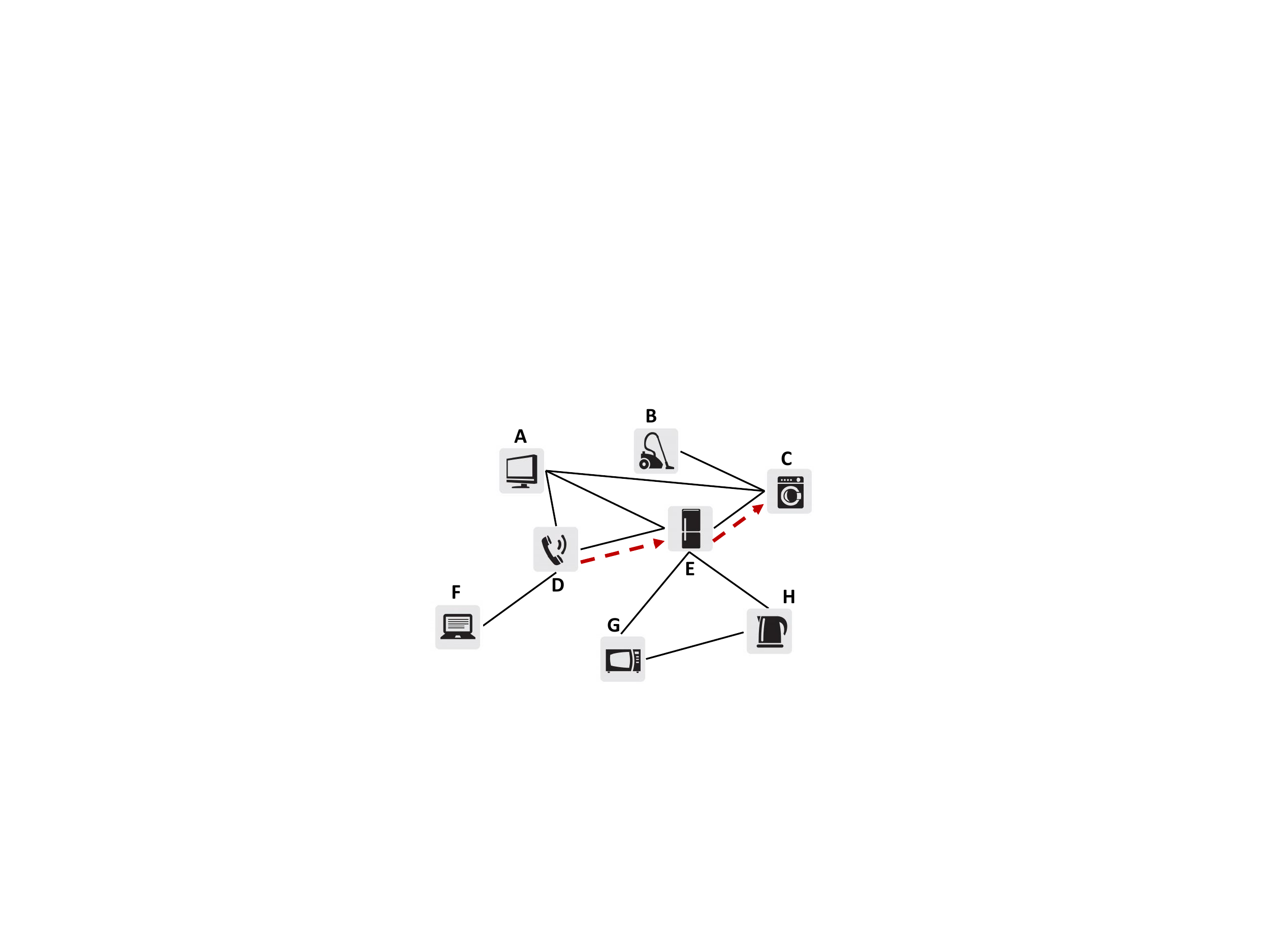}
\caption{SAND in a Smart Home}
\vspace{-1.5em}
\label{sand}
\end{figure}

An illustrative example of SAND in a smart home is shown in Fig. \ref{sand}. Here, we only show the overlay device social network. We assume that all the devices are connected through WiFi, so the communication network is a complete mesh network, which is not shown. In the overlay social network, the refrigerator ($E$), the TV ($A$) and the washing machine ($C$) are connected since they are from the same manufacturer, e.g., Samsung. The neighbor relationship exists between the vacuum ($B$) and the washing machine ($C$), both of which locate in the storage room. Friendship exists between (1) the telephone ($D$) and the TV ($A$), and (2) the refrigerator ($E$) and the PC ($F$), which are involved in frequent interactions to serve the home owner. The refrigerator ($E$), the microwave ($G$) and the boiler ($H$) are considered as colleagues that work together to prepare meals.

In SAND, when an IoT device joins the network, it will exchange the general information with nearby devices that are within the transmission range. Consequently, social links can be established if valid social relationships exist. Here, we assume each IoT device is social-aware and intelligent in the sense that they can establish social relationships autonomously like human. The social aspects of IoT devices give them the ability to form an overlay device social network automatically. It is worth noting that the overlay device social network is not static but dynamically changing because of the device movements or the device relationship changes (e.g., the colleague relationship may change frequently). Hence, each IoT device needs to periodically update their social connections in SAND.

In SAND, IoT devices are supposed to act as human beings. The social aspects of the devices allow them to dynamically create new social connections and form new colleagueship to work together to generate new services. With the social aspects in SAND, devices become more visible and aware to each other, thus leading to a fast and effective device discovery.

In the process of device discovery, rather than simply broadcasting the discovery message to all the neighbors as the distribute schemes does, SAND will send the discovery message to each of the neighbors in a preferred order based on the the rank of the neighboring device. The higher rank the device is, the more likely and faster it is able to forward the discovery message to the target device requested by the source device. With the social aspects in SAND, devices become more visible to other devices, thus making SAND more scalable and efficient than the baseline distributed scheme. 

\subsection{The Device Ranking Criteria in SAND}
In SAND, the device discovery adopts the depth-first search strategy, where the discovery request is forwarded to the neighboring devices in a preferred order based on their ranks. Here, the rank of a device is determined by four factors which are the device degree, the diversity, the (local) clustering coefficient and the (local) betweenness of the IoT device. The first three factors help SAND to select a device that may reach a broad community, while the betweenness leads the discovery request to a device that stands at the major intersection of multiple shortest paths in the network. Thus, the device ranking criteria makes SAND an intelligent, accurate and fast device discovery strategy. We will present the calculation details of these four factors as follows.

\emph{Device Degree:} In the overlay social network, the degree of device $i$ is denoted by $k_i$, which is defined as the number of social links it has. For example, in Fig. \ref{sand}, device $E$ has a degree of 5 since there are five social connections associated with it. The higher the device degree is, the more likely it is associated with a routing path to the desired target device since there are more outlets from this device.

\emph{Diversity:} In SAND, the diversity of device $i$ is denoted by $d_i$, which is defined as the number of types of social links a device is associated with. For example, in Fig. \ref{sand}, device $E$ has a diversity of 3 since it has three different types of social links which are family (i.e., links with $A$ and $C$), colleagueship (i.e., links with $G$ and $H$) and friendship (i.e., link with $F$).  An IoT device could have a high device degree $k_i$, but if all the connections are within the same type of social relationship, the device is still assigned with a relatively low rank since it is not diverse enough to reach distinct communities of devices. In contrast, an IoT device that has a high diversity $d$ is involved in many different types of social relations, and may have a broad reachability, thus having a higher chance to connect with the desired target device.

\emph{Clustering coefficient:} The clustering coefficient shows how likely a device and its neighbors form a clique \cite{watts}. The local clustering coefficient of a device is defined as the number of the social links in its neighborhood (including the device itself and its one-hop neighbors) divided by the total number of possible links in the neighborhood. For a given device $i$, the local clustering coefficient $c_i$ can be calculated as:
\begin{equation}
c_i = \dfrac{2|\big\{e_{st}:s, t\in N_i, e_{st}\in E\big\}|}{k_i(k_i-1)}
\end{equation}
where $e_{st}$ is a social link between device $s$ and device $t$, $N_i$ is the set of devices in the neighborhood of device $i$, $E$ is the set of social links in the neighborhood of device $i$, and $k_i$ is the degree of device $i$. The higher the local clustering coefficient is, the more likely the device and its neighbors are forming a clique (e.g., fully mesh), thus the diameter of the device neighborhood is smaller, which can lead to a faster and wider dissemination for the discovery request. 

\emph{Betweenness:} The betweenness reflects the probability that a given device stands at the critical intersection of multiple shortest paths in the network \cite{newman}. It is defined as the number of shortest paths that traverse the given device divided by the total number of possible shortest paths. In SAND, the local betweenness measures such a probability in the neighborhood that includes the given device and its one-hop neighbors. The local betweenness of device $i$ is calculated as:
\begin{equation}
b_i = \dfrac{\sum_{s, t\in N_i}\delta_{st}(i)}{k_i(k_i-1)}
\end{equation}
where $N_i$ is the set of devices in the neighborhood of device $i$, $s$ and $t$ are a pair of devices in $N_i$, and $k_i$ is the degree of device $i$. Here, $\delta_{st}(i)$ is 1 if the shortest path between $s$ and $t$ traverses through device $i$; otherwise, it is 0. 
The higher the local betweenness is, the higher chance the given device is located at the intersection that connects major shortest paths in the neighborhood. Discovery requests that arrive at such intersection can be disseminated to anyplace in the network easily and quickly over those shortest paths.

In SAND, the rank of a device $R_i$ is defined as the multiplication of the above mentioned four factors (e.g., Eq. (3)). The higher the rank is, the more likely the device can forward the discovery request to the desired target device in a fast and efficient manner.
\begin{equation}
R_i = k_i * d_i * c_i * b_i
\end{equation}

\subsection{The SAND Forwarding Strategy}

Based on the architecture and the device ranking criteria as described above, we propose the SAND device discovery scheme in this subsection. 
Rather than the simple broadcast used in the distributed scheme, SAND forwards the discovery request to a neighbor device that ranks the highest. For example, in Fig. \ref{sand}, the discovery request from $D$ to $C$ will be forwarded to $E$, which has the highest rank (with a degree of 5, diversity of 3, local clustering coefficient of 0.53 and local betweenness of 0.53). The device with the highest rank is supposed to be the most effective one that leads to the desired target device. 
In the discovery process, SAND performs a depth-first search with limited search depth, where the search will not exhaust to the deepest level but a depth level of $n$ (which is a tunable parameter). If the desired target device cannot be found after searching for $n$ levels, SAND will step back to examine the other unchecked neighbors in depth $n-1$. 
The pseudocode of SAND discovery algorithm is shown in Algorithm 1.

\begin{algorithm}
\caption{The SAND Device Discovery Algorithm}\label{SAND}
\begin{algorithmic}[1]
\renewcommand{\algorithmicrequire}{\textbf{Input:}}
\renewcommand{\algorithmicensure}{\textbf{Output:}}
\Require A discovery request $r_{st}$ from source $s$ to target $t$
\Ensure The communication path $p_{st}$ between $s$ and $t$
	\State $p_{st}.push(s)$;
	\While{$p_{st}$ is not empty}
		\State $i \leftarrow p_{st}.pop()$;
		\If{$i$ is the desired target device}
			\State return the reverse path of $p_{st}$;
		\ElsIf{$i$ has been checked before}
			\State continue;
		\ElsIf{$p_{st}.length()$ = depth $n$}
			\State continue;
		\Else
			\State rank and sort the neighbor devices $N_i$;
			\State $p_{st}.push(N_{i\_highest\_rank})$;
		\EndIf
	\EndWhile
\end{algorithmic}
\end{algorithm}

Similar as the distributed scheme, SAND is scalable since each IoT device only maintains information of its local connections. SAND is also resilient and robust since there is no centralized controller and each IoT device is fair and equal as a peer. Compared to the distributed scheme, SAND is more efficient because the former applies simple broadcast and thus a significant large number of devices are involved in the discovery process, while in the latter, SAND performs depth-first search with limited search depth and the guidance of an intelligent device ranking criteria, so the number of devices involved is relatively small. Such an advantage of SAND may lead to a low energy consumption in the process of transmitting the discovery request, which is critically beneficiary to the IoT system where devices are power constrained. Hence, we can see that SAND inherits the advantages of the distributed scheme, and further improves the communication efficiency.

\section{Performance Evaluation}

In this section, we focus on evaluating the performance of the proposed SAND scheme, compared to the distributed scheme that uses simple broadcast. We conduct simulations on both a random network and a scale-free network. 
The random network is an irregular mesh network with social links randomly generated between IoT devices, while the scale-free network is generated by considering the power law \cite{barabasi} in which the fraction of devices with $k$ connections follows a distribution $P(K) \sim k^{-\gamma}$ ($\gamma$ is usually between 2 and 3). 
It is meaningful and reasonable to test SAND on a scale-free network, since most of the social networks are scale-free and SAND works by considering an overlay device social network.

In our simulation, we generate a network that consists of 20000 IoT devices. Each IoT device is associated with a number of connections to its neighboring devices (the number of connections follows uniform distribution and power law in the random network and the scale-free network, respectively), which forms the communication network. 
Each physical link in the communication network has a transmission latency uniformly distributed within [10ms, 40ms]. 
The distributed scheme only runs on this communication network. In SAND, for each IoT device, we randomly generate three features. Any two IoT devices that are physically connected and share the same features will be equipped with a social link, which forms the overlay device social network. The total number of features in the network is set to vary from 2000 to 10000. For all the simulation results shown below, we generate 15000 discovery requests from randomly selected source devices with randomly chosen desired features, and obtain the average results. Each discovery request has a time to live of 60s.
With regards to the evaluation, we focus on three performance metrics, which are the success rate, the average number of devices contacted on success and the average number of hops of the discovery path on success. 
We will present the simulation results and findings in the following parts.

\begin{figure*}[!t]
\centering 
\subfloat[Success Rate in Random Networks]{\includegraphics[width=3.2in]{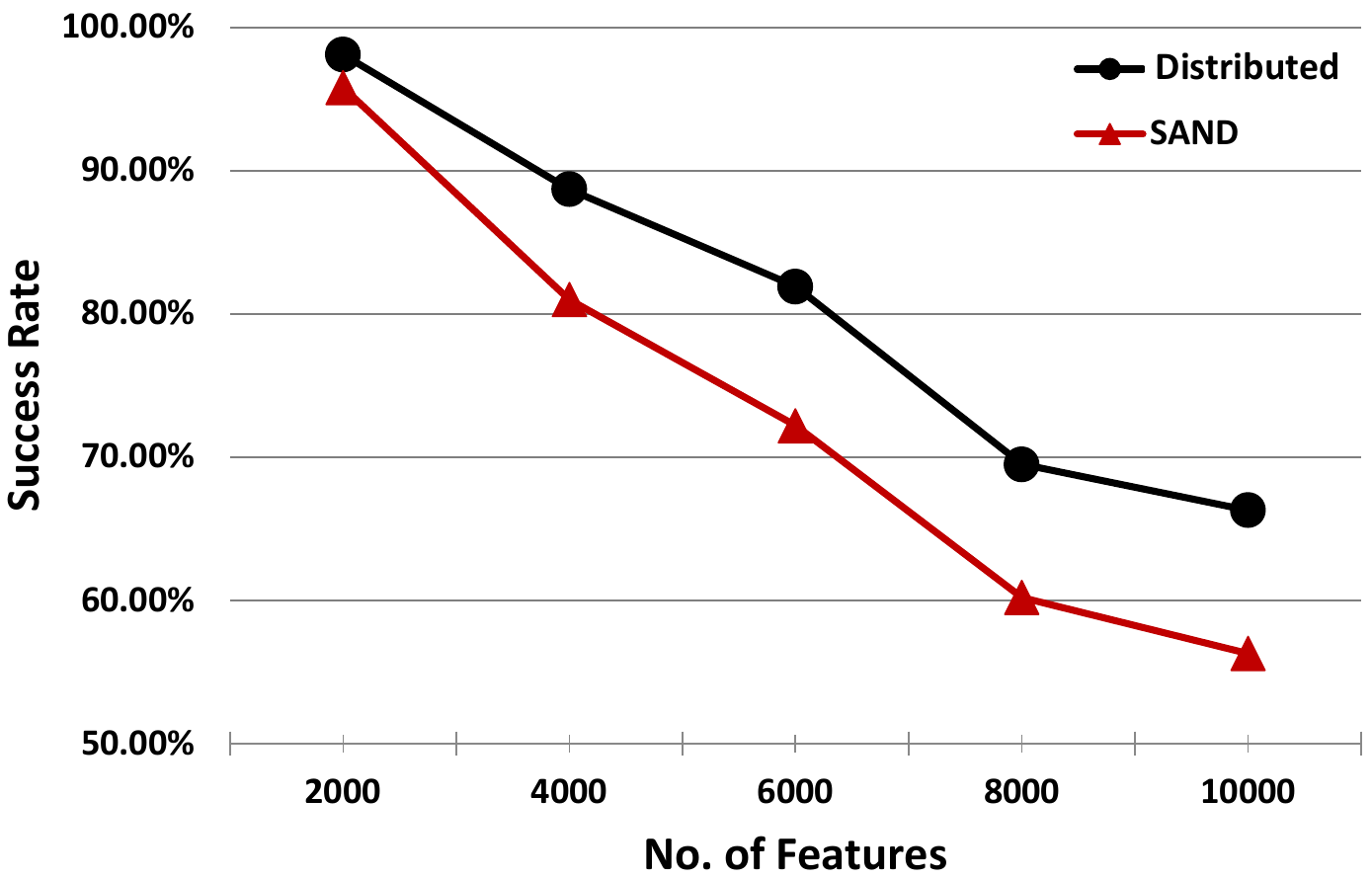}}\quad 
\subfloat[Success Rate in Scale-free Networks]{\includegraphics[width=3.2in]{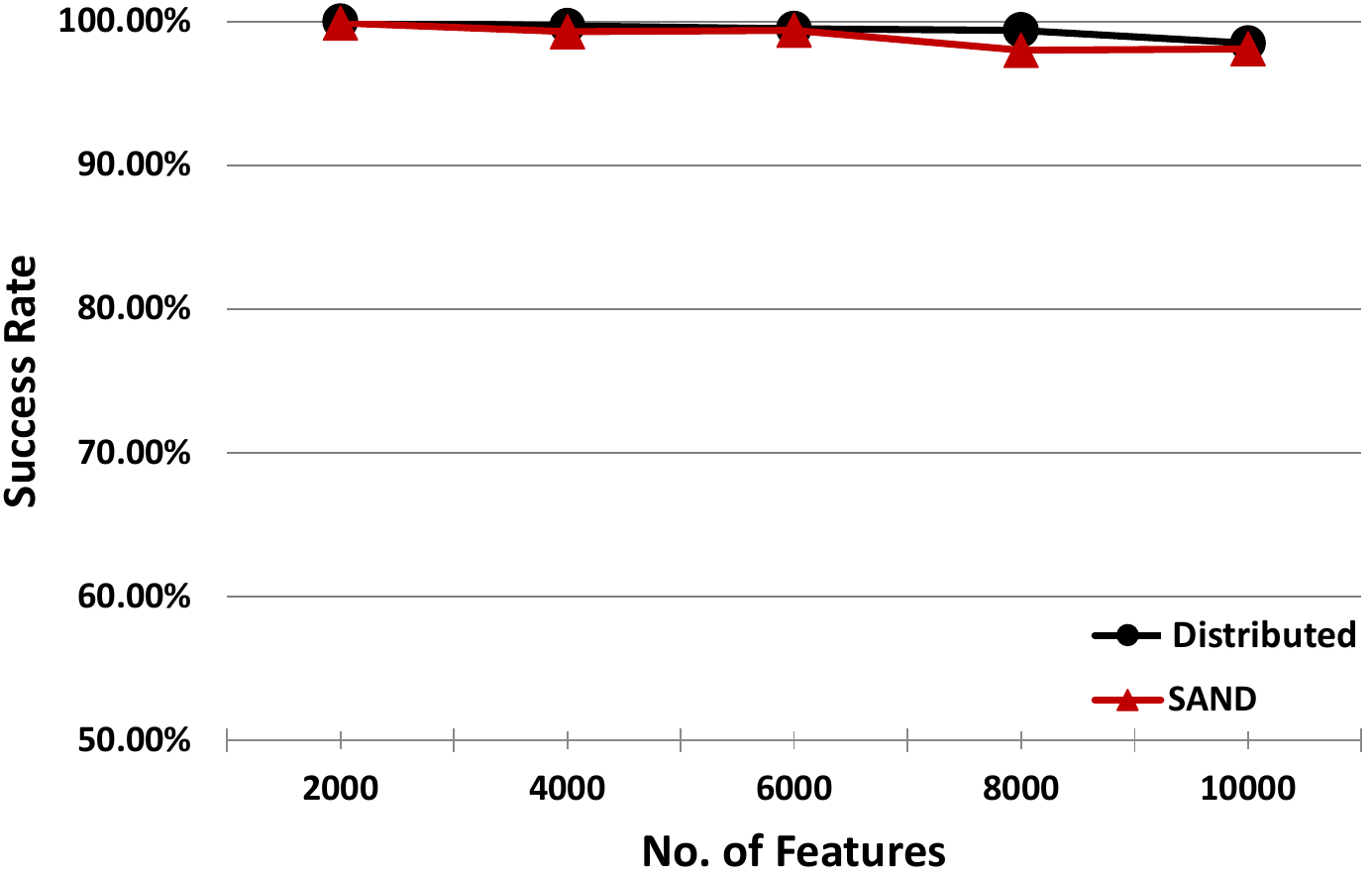}}\\ 
\caption{Simulation Results in the Random Network and the Scale-free Network} 
\vspace{-1.5em}
\label{sim}
\end{figure*}

\subsection{Success Rate}

In the simulation, if the target device with the desired feature can be found before the time to live expires, it is considered as a successful discovery; otherwise, it is considered as a failure and the discovery request will be dropped.
The success rate is defined as the number of successful discovery divided by the total number of discovery requests. We show the success rate as a function of the number of features in Fig. \ref{sim}(a) and Fig. \ref{sim}(b). Given a fixed number of IoT devices (i.e., 15000 by default) in the system, as the number of features increases, the system become more diverse with a variety of types of devices. From Fig. \ref{sim}(a), we can see that in the random network the success rate of both the distributed scheme and SAND decreases as the network becomes more diverse. The distributed scheme performs better than SAND since the distributed scheme uses simple broadcast while SAND uses the depth-first search which may yield to a long discovery period that exceeds the required time to live. 

However, in the scale-free network as shown in Fig. \ref{sim}(b), we can see that SAND achieves a similar success rate as that of the distributed scheme, which is as high as more than 98\%. Compared to the results in the random network, the success rate experiences a significant increase. This is primarily because in the scale-free network there are some superhub devices that have a large number of connections, which could help the discovery request reach the target device in a short time. While in the random network, each device has a similar node degree and the discovery request may take a long time to reach the target, and in some bad cases the time to live requirement is violated and thus failures occur. Another reason is that the device ranking criteria in SAND is effective and can successfully navigate the discovery request to the desired target device. Given that most of the real-world social networks are scale-free, we can infer that SAND is a practically solid solution since its success rate is above 98\%.

\begin{figure*}[!t]
\centering 
\subfloat[Avg. No. of Devices Contacted in Random Networks]{\includegraphics[width=3.2in]{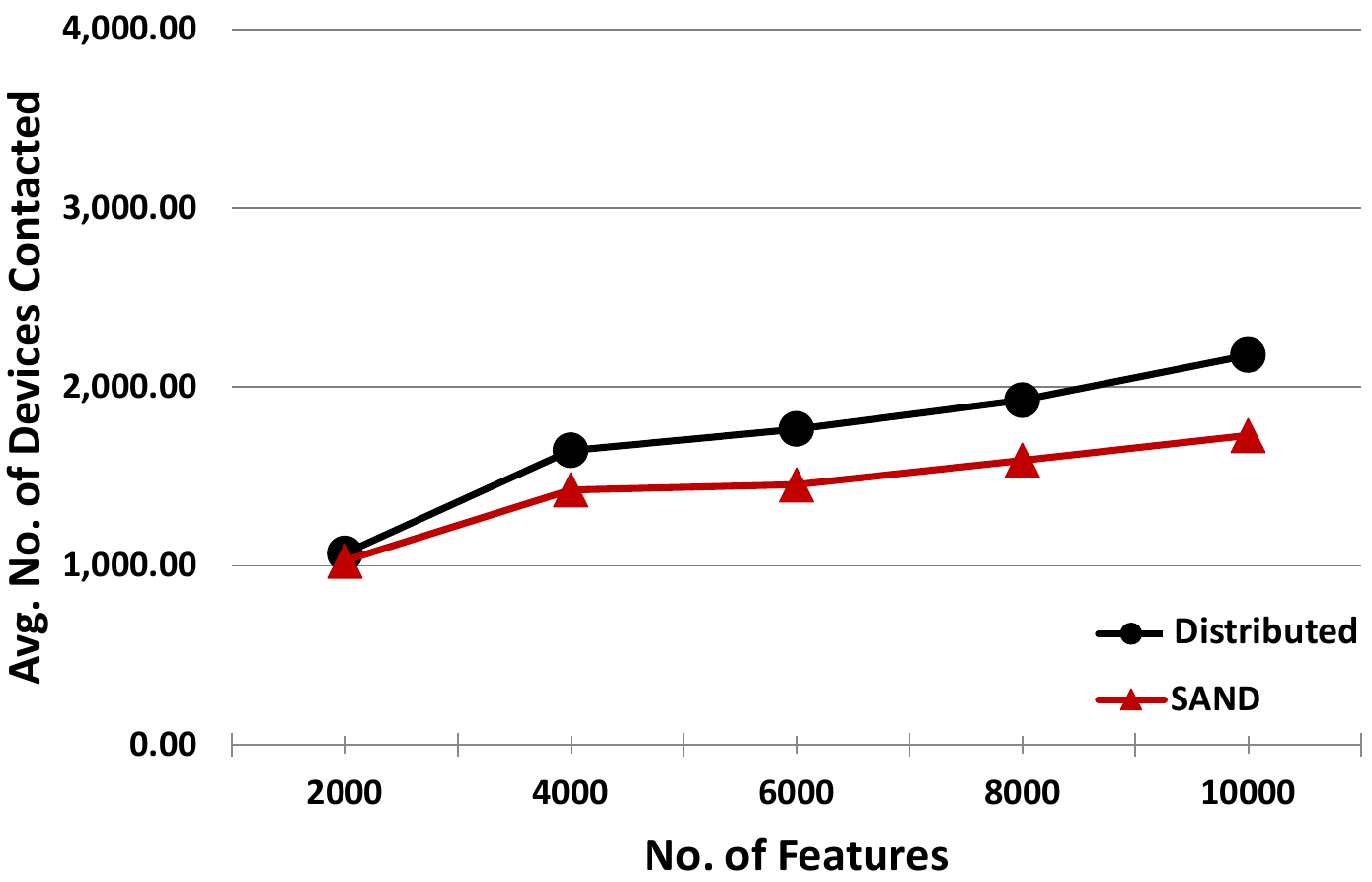}}\quad 
\subfloat[Avg. No. of Devices Contacted in Scale-free Networks]{\includegraphics[width=3.2in]{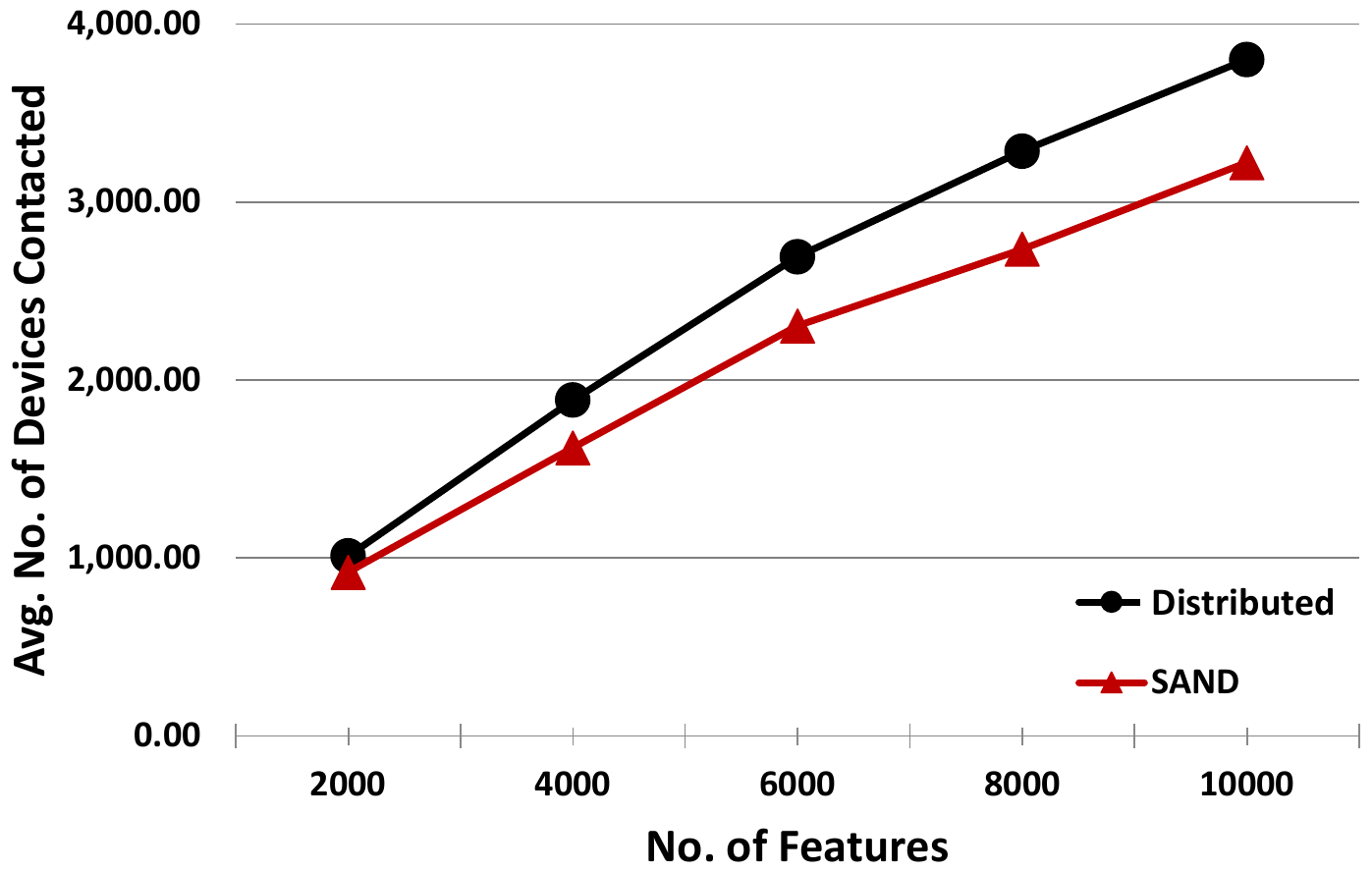}}\\ 
\caption{Simulation Results in the Random Network and the Scale-free Network} 
\vspace{-1.5em}
\label{sim2}
\end{figure*}

\subsection{Average Number of Devices Contacted}

One of our main interests in this research is to measure, during a device discovery process, how many devices are being contacted before the desired target device is found. 
It is more energy efficient if less number of devices are involved in transmitting the discovery message, which is critically important for IoT system where devices are power constrained. 
Fig. \ref{sim2}(b) and Fig. \ref{sim2}(b) show the number of devices contacted with different number of features in the random network and the scale-free network, respectively. Here, we only count the measures on successful discovery.
From the simulation results, we can observe that in both random and scale-free networks, as the number of features increases, SAND outperforms the distributed scheme, with an average performance improvement of 14.6\% and 13.9\%, respectively. It can also be seen that as the number of features increases, the performance improvement of SAND over the distributed scheme becomes more significant in both networks. This indicates that SAND is more efficient when the IoT system is more diverse and heterogeneous. 
Thanks to the adoption of depth-first search and the intelligent device ranking criteria, SAND can efficiently discover the desired target device without involving too many unnecessary intermediate devices. Thus, SAND can potentially achieve a low energy consumption in the process of transmitting the discovery request, which is critically beneficiary to the IoT where the devices are power constrained.

\subsection{Communication Hops}

Another interest in this research is to find out how many communication hops are there to separate a source device from a desired target device. After the discovery process is done, the source and target devices will communicate with each other over the discovery path and cooperate to perform computing tasks. The post-discovery communication can be cost-efficient if the number of hops of the discovery path is small. In Table I and II, we show the average number of hops of the discovery path in the random network and the scale-free network, respectively. It can be seen that SAND can achieve as small communication hops as that of the distributed scheme, the latter of which is supposed to be the optimal one because of using the breadth-first search. This finding further validates the effectiveness of the device ranking criteria of SAND, which can intelligently navigate the discovery to the desired target device with minimum detours. We can also see that as the number of features increases, the number of hops increases in both network scenarios. This is reasonable because it becomes more difficult and needs more hops of query to find the desired target device when the network becomes more diverse. Comparing the results in the two tables, we can also observe that from the random network to the scale-free network, the average number of hops exhibits an average increase of two hops. The reason behind is that the random network is a regular mesh network where all the devices are relatively fair and have similar number of connections, while the scale-free network has some superhub devices that have huge number of connections. In the scale-free network, a given discovery request is usually forwarded to those superhub devices first and then finds a path to the desired target device, thus resulting in a larger number of hops than that of the random network.

\begin{table}
\centering
\caption{No. of Hops in Random Networks}
\label{tab:1}       
\begin{tabular}{llllll}
\hline\noalign{\smallskip}
No. of Features & 2000 & 4000 & 6000 & 8000 & 10000 \\
\noalign{\smallskip}\hline\noalign{\smallskip}
Broadcast & 3.67 & 4.16 & 4.43 & 4.53 & 4.82 \\
\noalign{\smallskip}\hline\noalign{\smallskip}
SAND & 3.66 & 4.14 & 4.28 & 4.40 & 4.69 \\
\noalign{\smallskip}\hline
\end{tabular}
\end{table}

\begin{table}
\centering
\caption{No. of Hops in Scale-free Networks}
\label{tab:1}       
\begin{tabular}{llllll}
\hline\noalign{\smallskip}
No. of Features & 2000 & 4000 & 6000 & 8000 & 10000 \\
\noalign{\smallskip}\hline\noalign{\smallskip}
Broadcast & 5.41 & 6.15 & 6.37 & 6.47 & 6.70 \\
\noalign{\smallskip}\hline\noalign{\smallskip}
SAND & 5.41 & 6.14 & 6.36 & 6.41 & 6.69 \\
\noalign{\smallskip}\hline
\end{tabular}
\end{table}

\begin{figure}[!t]
\centering
\includegraphics[width=3.2in]{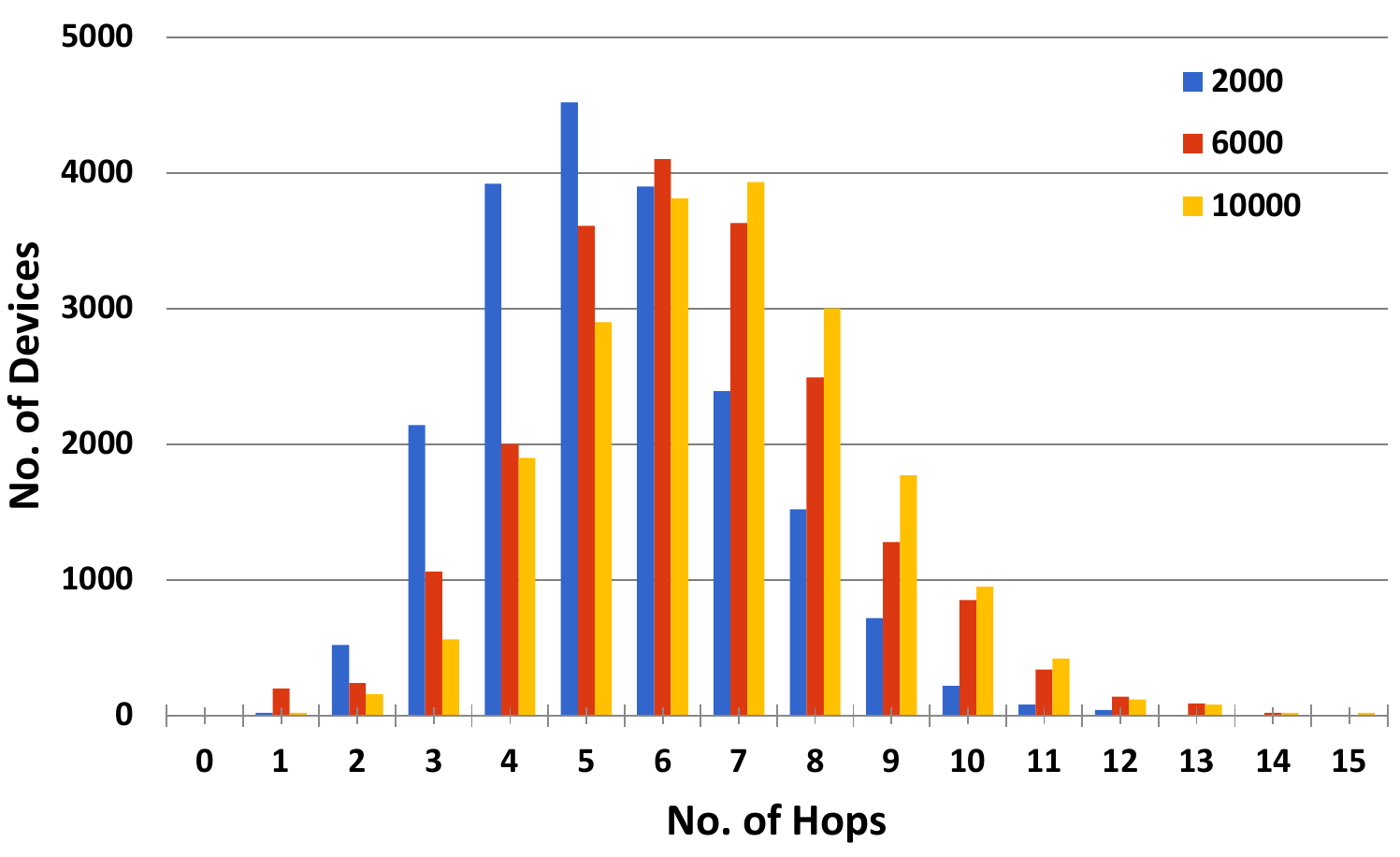}
\caption{Device Distribution v.s. No. of Hops in SAND with 20K nodes}
\vspace{-1.5em}
\label{hop}
\end{figure}

Furthermore, we evaluate the distribution of devices over the number of hops of the discovery path in the scale-free network. As shown in Fig. \ref{hop}, we plot the device distributions for SAND using 2000, 6000 and 10000 features. The result confirms the past research on social networks \cite{travers}\cite{fb}, and shows that in most of the cases any two IoT devices are separated by 5$\sim$7 hops.  We can also observe that the average number of hops is small when the network is less diverse (see the center of the peak of distribution with 2000 features), and the average number of hops increases when the network becomes more diverse (e.g., 6000 and 10000 features). We can also infer that if a device cannot be discovered in a few hops, the probability that the discovery is unsuccessful increases dramatically. The reason behind is that the unsuccessful discovery involves devices that are relatively isolated from all other devices. It is quite common that the discovery to those isolated devices will violate the time to live requirement and get dropped. In contrast, the successful discovery usually involves devices that are in a cluster or connected to the superhubs.

\section{Research Challenges and Opportunities}
An efficient IoT device discovery may lead to the generation of a new set of applications and services, as well as research opportunities, such as self-organized computing, service and data fusion, malware source traceback and the formation of a robot society, etc. Meanwhile, there are many open challenges with regards to these applications. In the following parts, we propose a few research opportunities and challenges related to the topic of intelligent device discovery in the IoT, and beyond.

\subsection{Self-Organized Computing}
The intelligent device discovery can enable a new format of computing services that can be self-organized by a number of smart devices. A client can initiate a computing task from any device (e.g., a smart watch), with a set of specified requirements such as computing power demand (e.g., CPU and memory), function or platform demand, and data demand (e.g., source code or raw dataset). The source IoT device is responsible for deliverying these requirements to the appropriate computing facilities in the Internet and gathering the results back. In this case, an intelligent device discovery strategy can be used to establish connections between IoT devices and the computing facilities in the Internet. For example, a user wants to issue a request to check his/her glucose level from the smart watch. The smart watch can use the intelligent device discovery to search for the nearest server that has enough computing and memory power to perform the computing tasks. Then, the server may use the intelligent device discovery to find the source code of glucose level analyzer, and then download and install the analyzer to perform the analysis. The intelligent device discovery grants IoT devices and computing facilities the capability to fetch the data and functions as they need for performing a given computing tasks. Thus, the IoT devices and computing facilities can perform the computing tasks in a self-organized manner. 

In this process, one of the open challenges is how to standardize the experssion of the computing requirements so that the devices and machines can initiate standard requests that can be understood by other machines. Another challenge is how to optimize the request and reply flow routes so that the network resource consumption can be minimized and the request-response latency can be minimized.

\subsection{Computing, Network and Data Fusion}
In the above self-organized computing scenario, a variety of computing facilities, network functions/services and data (e.g., programming source code, patient's raw data) can be interconnected and fused by running an intelligent device discovery strategy. For a given service request, the intelligent device discovery can help with gathering the computing, switching and storage hardware from a specific site, while fetching a pool of patients' data from a different site and obtaining the source codes of particular computing functions or network functions from another site. Here, the computing, switching and storage hardware are served as placeholders, while the source codes of the computing and network functions and the raw dataset can be discovered and fetched by the intelligent device discovery strategy and served as the input to those hardware placeholders. Data, software and hardware can be distributed in different sites, while they are coupled by the intelligent device discovery approach. 
Hence, an intelligent device discovery strategy is promising to enable the computing, network and data fusion, and achieve an efficient communication between the fused devices.
After the computing, network and data fusion becomes true, devices and machines can easily launch new services by specifying their needs to the Internet. The Internet can resolve the needs in a standard format (e.g., service function chains) and efficiently connect those required devices and funcitons (e.g., through intelligent device discovery strategy) to provide the required new services specified by the devices and machines. 
With the computing, network and data fusion, it is promising to generate novel machine-initiative services that are not supervised by human beings.
Fusion can be considered as an intermediate phase that prepares and enables the formation of a robot society.

To achieve this, an open challenge is how to develop a platform that can address the IoT interoperability issue and facilitate a computing, network service and data fusion platform. Another challenge is how to efficiently discover and deliver the required data and source codes to the selected hardware sites in a timely manner such that a seamless service can be provided for the users.

\subsection{Traceback Cyber Attack Source}
Another application of intelligent device discovery is in the area of network security. The sources of malicious attacks are usually spoofed to avoid being detected \cite{traceback}. The intelligent device discovery can be used to traceback the actual source of malware attacks based on the characteristics and features of the malicious attacks. The security protection system can issue a traceback request for the malware source generator, with certain key features of the malicious traffic. The overlay social network on top of the communication network can be constructed based on the model in Section 3.1. Then, an enhanced model that can accurately predict the hidden links/relationships between machines can be constructed using the method in \cite{predict}. 
Machine learning or deep learning \cite{chang3} techniques can also be applied here to improve the accruacy of the prediction.
After that, the proposed intelligent device discovery strategy can run on this enhanced overlay social network (with hidden links based on prediction) to traceback the malicious source attacker. Even if the malicious source are spoofed and hidden behind some links or relationships, the intelligent device discovery can still guide the traceback route based on the pattern and features of malicious attacks.

\subsection{Robot Society}
The intelligent device discovery can be used to initiate the connection between a given device and its desired device for the first time. After the first contact, the two devices can be considered to know each other and maintain a certain type of social relationship as human does. As the communications between different devices continue to grow, a variety of computing service, network service and data can be fused, and a robot society could be formed which consists of IoT devices, servers, robots and all the machines in the Internet. The device relationship in the robot society can be the same as the human relationship in the human society, such as friend, colleague and family relationships (which are considered in this paper). There is also a high chance that some unique machine type of relationships may exist in the robot society. One of the challenging and meaningful work is to analyze a large dataset of network traffic data to categorize the relationships between machines/devices in the Internet. This can be useful to give us an insight of the relationships and communication purpose between machines. It is also helpful for us to formulate the foundamental rules of how to establish the social relationship between machines/devices. Another open challenge is how to ensure the security of the connection before the two machines/devices establish their social relationship. It is necessary to avoid establishing relationships with potential malicious devices. To achieve this, we may need to develop machine learning based solutions (e.g., decision tree classification, convolutional neural networks or recurrent neural networks) to train the machines/devices to identify their own roles and their contact's roles in the robot society,  to reduce the chance of establishing connections with malicious devices. In general, a promising research area is to investigate the social behavior and the collaborative intelligence in the robot society, in addition to the artificial intelligence in a single device/machine.

\section{Conclusion and Future Work}

The Internet of Things (IoT) grows continuously and connects billions of smart devices with heterogeneous functionalities, purposes and platforms. 
As the IoT evolves, it is promising that human-like social relationships could be autonomously established by the smart devices. With the help of an intelligent device discovery strategy, IoT devices can easily discover the resources (e.g., source codes) and devices (e.g., computing facilities) that meet their needs. Each device/machine is associated with one or more resources or capabilities of either computing power, network function, source code or data. The intelligent device discovery can be used to glue and couple the resources from distributed sites to achieve the computing, network service and data fusion. Each device/machine acts like a robot and can establish social connections with each other autonomously to collaborate and create new applications and services, which could eventually lead to the formation of a robot society.
The goal of this research is to address the impending scalability issue in the process of device discovery in IoT. In this paper, we have proposed a novel approach that leverages such social aspects of IoT devices to achieve a scalable and efficient IoT device discovery. The proposed \emph{Social-Aware and Distributed} (SAND) scheme applies the depth-first search and forwards the discovery request to neighboring devices in a preferred order defined by a novel social network-aware device ranking criteria. More specifically, the ranking criteria takes into consideration the device's degree, diversity and clustering coefficient, which could potentially navigate the discovery request to reach a broad community. Furthermore, the local betweenness is considered in the ranking criteria, which prioritizes the devices that stands at the critical intersections of multiple shortest paths of the network. 
With the guidance of such an intelligent device ranking criteria, the discovery request can find the desired target device in a fast and efficient manner. 
We have conducted comprehensive simulations to validate the effectiveness of SAND on both a random network and a scale-free network. Simulation results have shown that SAND can achieve the near-optimal success rate and communication hops as that of the distributed scheme which uses broadcast. In addition, we have found that the SAND scheme contacts a much smaller number of intermediate devices than that of the distributed scheme during the discovery process, thus potentially leading to a much less energy consumption. 

As for future work, we plan to extend our study in the following aspects. First, we plan to extend SAND with multicast feature, which allows one discovery request to fetch multiple replies. This could be useful in wireless sensor network applications, such as in the temperature monitor system. In addition, rather than specifying only one feature of the desired target device, we can enhance the SAND scheme by allowing it to claim for multiple features to achieve a more accurate discovery. We also plan to implement SAND on a physical testbed and conduct experimental studies. Secondly, we plan to analyze various network traffic datasets to understand the meanings and purposes behind those communications, and we can use machine learning techniques to categorize the social relationships in the IoT. This can be used to further improve the discovery speed and the discovery accuracy of SAND, and can also provide the basic insights for exploring the formation of the robot society. Thirdly, we plan to investigate the interoperability issues in the IoT and explore a standard platform that enables a variety of computing, network service and data fusion.






%

\end{document}